\documentclass{article}

\PassOptionsToPackage{numbers, compress}{natbib}

 \usepackage[final]{neurips_2025}

\usepackage[utf8]{inputenc} 
\usepackage[T1]{fontenc}    
\usepackage{hyperref}       
\usepackage{url}            
\usepackage{booktabs}       
\usepackage{amsfonts}       
\usepackage{nicefrac}       
\usepackage{microtype}      
\usepackage{xcolor}         
\usepackage{graphicx}
\usepackage{booktabs}
\usepackage[misc]{ifsym}
\usepackage{amsmath}
\usepackage{natbib}

\title{Inductive Transfer Learning for Graph-Based Recommenders}

\author{%
  Florian Grötschla\thanks{Equal contribution.} \quad
  Elia Trachsel\footnotemark[1] \quad
  Luca A. Lanzendörfer \quad
  Roger Wattenhofer \vspace{0.5cm}\\
  ETH Zurich \\
  \texttt{\{fgroetschla, trachsele, lanzendoerfer, wattenhofer\}@ethz.ch}
}

\begin{document}

\maketitle

\begin{abstract}
  Graph-based recommender systems are commonly trained in transductive settings, which limits their applicability to new users, items, or datasets. We propose NBF-Rec, a graph-based recommendation model that supports inductive transfer learning across datasets with disjoint user and item sets. 
Unlike conventional embedding-based methods that require retraining for each domain, NBF-Rec computes node embeddings dynamically at inference time. We evaluate the method on seven real-world datasets spanning movies, music, e-commerce, and location check-ins. NBF-Rec achieves competitive performance in zero-shot settings, where no target domain data is used for training, and demonstrates further improvements through lightweight fine-tuning. These results show that inductive transfer is feasible in graph-based recommendation and that interaction-level message passing supports generalization across datasets without requiring aligned users or items.
\end{abstract}

\section{Introduction}

Recommender systems play a central role in many digital platforms, where they personalize content by modeling interactions between users and items. Graph-based approaches have proven effective in this setting, as they capture complex relational dependencies through message passing on user-item interaction graphs. However, most graph-based recommenders are trained and evaluated within a single dataset, making strong assumptions about the availability of domain-specific interaction histories. This limits their ability to generalize to new users, items, or domains, and restricts their applicability in real-world scenarios where data is sparse or fragmented across platforms.
In contrast, transfer learning has enabled substantial progress in fields such as natural language processing and computer vision, where models pretrained on large datasets can be adapted to new tasks with limited supervision. Despite this success, transfer learning remains underexplored in the context of graph-based recommender systems. Existing methods often rely on transductive training paradigms or assume overlapping user or item sets across domains. These constraints hinder their ability to perform recommendation in settings where no such overlap exists and where target domain data is unavailable during training.
We address this gap by introducing NBF-Rec, a graph-based recommendation model that supports inductive transfer learning across disjoint user-item graphs. NBF-Rec computes node representations dynamically via message passing, without relying on pretrained embeddings or access to the target domain during training. This design enables zero-shot recommendation in new domains and supports further adaptation through fine-tuning when limited target interactions are available.
We evaluate NBF-Rec on seven real-world datasets spanning diverse application areas, including movies, music, e-commerce, and location-based services. Our experiments show that the model achieves competitive performance in zero-shot scenarios, and that fine-tuning on a small subset of target interactions leads to further improvements. These results demonstrate the viability of inductive transfer learning in graph-based recommendation.

\begin{figure}
    \centering
    \includegraphics[width=0.7\linewidth]{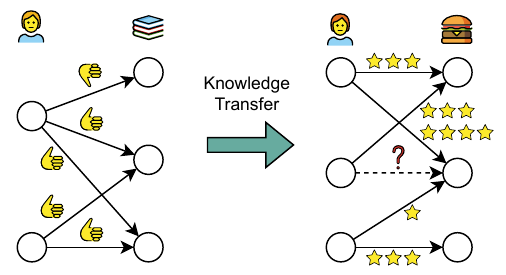}
    \caption{Schematic overview of NBF-Rec’s inductive transfer learning pipeline. The model is trained on user-item interaction graphs from one or more source domains and applied to unseen target domains. Transferable patterns are learned from structural and interaction features without requiring shared users or items.}
    \label{fig:enter-label}
\end{figure}

\section{Related Work}
Graph Neural Networks (GNNs) have become central in recommendation systems due to their ability to model complex user-item relationships. Early models like LightGCN~\cite{he2020lightgcn} achieved strong performance by simplifying message passing, but were transductive in nature and failed to generalize to unseen nodes or graphs. To enable inductive reasoning, models such as MINERVA~\cite{das2017go}, GraIL~\cite{teru2020inductive}, and RED-GNN~\cite{zhang2022knowledge} introduced path- and subgraph-based methods. NBFNet~\cite{zhu2021neural}, which forms the foundation of our approach, learns to aggregate path-based messages in a query-dependent manner, supporting inductive link prediction.
More recent work has explored transfer learning in GNNs to improve generalization across domains. ULTRA~\cite{galkin2024ultra} extended NBFNet for zero-shot knowledge graph completion by jointly reasoning over relation and entity graphs. In recommendation, most cross-domain models assume shared users or items~\cite{zhu2021cross}, limiting their applicability. Other works employ adversarial training~\cite{li2022recguru}, contrastive disentanglement~\cite{li2025disco, 10.1007/978-3-031-30672-3_11, liu2024graph, ren2023disentangled}, or meta-learning~\cite{vartak2017meta} to address cross-domain generalization, but often rely on aligned entity spaces or domain-specific supervision.
Efforts leveraging pretrained language or vision models, such as U-BERT~\cite{qiu2021ubert} and ChatRec~\cite{gao2023chatrec}, have shown promise for cross-domain recommendation but typically require rich side information. In contrast, NBF-Rec operates solely on interaction graphs, without depending on textual or visual content. Contrastive learning methods like SimGCL~\cite{yu2022simgcl} and NESCL~\cite{sun2023nescl} enhance transferability by regularizing representations, though they focus primarily on improving performance within a single domain.
Large-scale pretrained models have recently been explored as a unified solution to recommendation tasks. P5~\cite{geng2022recommendation} formulates personalized recommendation as a text-to-text problem and leverages large language models to handle multiple tasks across domains in a unified framework. Similarly, GPTRec~\cite{petrov2023generative} adapts generative pretrained transformers to model user behavior sequences for next-item prediction. While these models achieve strong performance in zero-shot and multitask settings, they require substantial pretraining and inference resources. In contrast, NBF-Rec provides a lightweight alternative based on graph message passing, enabling inductive transfer across datasets with disjoint user and item sets.
To our knowledge, no prior graph-based recommender has demonstrated scalable, inductive transfer across disjoint user-item graphs in a purely interaction-driven setting.

\section{Methodology}

\subsection{Problem Formulation}

Recommender systems can be formulated as a \textit{link prediction} task on a bipartite user-item graph. Let $\mathcal{U}$ be the set of users and $\mathcal{I}$ be the set of items. The history of interaction between users and items is represented as a graph $\mathcal{G} = (\mathcal{V}, \mathcal{E})$, where $\mathcal{V} = \mathcal{U} \cup \mathcal{I}$ is the set of nodes and $\mathcal{E} \subseteq \mathcal{U} \times \mathcal{I}$ denotes the interactions observed between users and items. Each edge $(u, i) \in \mathcal{E}$ represents an interaction (e.g., user buying a product, watching a movie, or listening to a song).
Given an incomplete graph $\mathcal{G}$, our goal is to predict missing edges, i.e., to recommend new items to users.

\subsection{NBF-Rec Model Architecture}

NBF-Rec is an inductive link prediction model that builds on neural Bellman-Ford networks (NBFNet) and integrates edge features into the message-passing process. Edge features capture rich interaction-level context and enable the model to distinguish different types and strengths of user-item interactions. For example, in the Epinions dataset, edge features include category, rating, helpfulness, and timestamp, which provide nuanced signals about user preferences and item relevance. Similarly, the LastFM dataset includes listening counts as edge features, reflecting users' interaction frequency and intensity with specific artists. Unlike traditional embedding-based approaches, which require retraining when new users or items appear, NBF-Rec dynamically computes node embeddings for previously unseen graphs.
Given a user \( u \), the message-passing procedure is defined as follows:
\paragraph{\textbf{Initialization.}}  
Each node \( v \) is initialized according to the standard NBFNet scheme:
$ h_v^{(0)} = \mathbf{1}(u = v),$
which ensures that all information propagates from the query user \( u \). Additionally, we define \( g \) as the function that generates edge embeddings from raw edge features \( r \):
\begin{equation}
    g(r) = \text{MLP}_{\text{emb}}\Big(\text{MLP}_{\text{proj}}(r)\Big).
\end{equation}
This function consists of a dataset-specific projection MLP followed by a backbone embedding MLP.

\paragraph{\textbf{Message Computation.}}  
At each layer \( t \), the messages for a node \( v \) are computed as:
\begin{equation}
    M_v^{(t)} = \Big\{ \text{MESSAGE}\Big( h_x^{(t-1)}, \mathbf{w}_q(x,r,v) \Big) \mid (x,r,v) \in \mathcal{E}(v) \Big\},
\end{equation}
where the edge weight is given by:
\begin{equation}
    \mathbf{w}_q(x,r,v) = \text{MLP}_t\big(g(r)\big).
\end{equation}
Here, edge features \( r \) are transformed into layer-specific embeddings via \( \text{MLP}_t \). The message function is implemented using a non-parametric \emph{DistMult} operation \cite{yang2014embedding}.

\paragraph{\textbf{Node Update.}}  
Each node \( v \) updates its representation by aggregating incoming messages along with its initial embedding:
\begin{equation}
    h_v^{(t)} = \text{UPDATE}\Big( h_v^{(t-1)},\, 
    \text{AGGREGATE}\Big( M_v^{(t)} \cup \{ h_v^{(0)} \} \Big) \Big).
\end{equation}
The aggregation function is a summation operation, followed by a learnable update function comprising a linear transformation, layer normalization, and an activation function.

\paragraph{\textbf{Score Generation.}}  
After \( T \) layers of message passing, we compute logit scores for each node \( v \) using a final MLP that projects the concatenation of the last-layer embedding and the initial embedding to a scalar value:
\begin{equation}
    \text{score}(u, q, v) = \text{MLP}_{\text{score}}\Big(\text{concat}\big(h_v^{(T)}, h_v^{(0)}\big)\Big).
\end{equation}
Notably, our model does not learn node-specific parameters, enabling node-inductive generalization. Unlike traditional embedding-based methods such as LightGCN \cite{he2020lightgcn}, which precompute node embeddings and apply simple similarity functions for recommendation, our approach dynamically performs message passing for each query.

\subsection{Training}  

Following the approach by~\citet{zhu2022torchdrug}, we train the model by minimizing a cross-entropy loss over positive and negative triplets:  
\begin{equation}  
\mathcal{L} = -\log p(u, q, v) - \sum_{i=1}^{n} \frac{1}{n} \log (1 - p(u'_i, q, v'_i)),  
\end{equation}  
where \( (u, q, v) \) is a positive interaction, and \( \{(u'_i, q, v'_i)\}_{i=1}^{n} \) are negative samples generated by corrupting either the head \( u \) or the tail \( v \). The negative samples are strict negatives, meaning they do not appear in the training dataset and are selected uniformly at random. This loss encourages the model to assign higher probabilities to positive interactions while minimizing the likelihood of negative ones. 
To optimize our model, we remove easy edges (i.e. batch edges) from the convolution graph during training. The training dataset defines the underlying graph for convolution, meaning that for every batch edge, the corresponding edge is already present in the convolution graph. If left unchanged, this setup would allow the model to trivially learn direct interactions rather than extracting meaningful relational patterns. By removing these edges, we force the model to rely on non-trivial paths.

\subsection{Computational Complexity}
NBF-Rec performs inference efficiently through dynamic message passing without precomputing user and item embeddings. A forward pass consists of (1) edge embedding transformation ($\mathcal{O}(|E|)$), (2) message passing ($\mathcal{O}(T|E|)$, with $T$ layers), and (3) score computation ($\mathcal{O}(|V|)$), resulting in overall linear complexity in the number of nodes and edges. While NBF-Rec incurs higher inference costs than models such as LightGCN, which precompute node embeddings, it enables transfer to unseen nodes, a capability absent in transductive architectures.

\section{Experimental Evaluation}
\begin{table}
    \centering
    \caption{Dataset statistics for the seven real-world benchmarks used in our evaluation. For each dataset, we report the number of users, items, and observed interactions, as well as the primary evaluation metric used in our benchmarking.}
    \begin{tabular}{lcccc}
        \toprule
        \textbf{Dataset} & \textbf{\#Users} & \textbf{\#Items} & \textbf{\#Interactions} & \textbf{Metric} \\
        \midrule
        ML-1M~\cite{harper2015movielens} & 5,950 & 2,811 & 364,654 & Hits@10\\
        LastFM~\cite{cantador2011second} & 1,867 & 1,867 & 39,717 & Hits@10\\
        Amazon B.~\cite{hou2024bridging} & 52,204 & 57,289 & 293,912 & Hits@10\\
        Gowalla~\cite{cho2011friendship} & 29,858 & 70,839 & 712,504 & NDCG@20\\
        Epinions~\cite{richardson2003trust} & 21,008 & 13,887 & 266,791 & Hits@10\\
        BookX~\cite{ziegler2005improving} & 12,720 & 18,318 & 276,334 & Hits@10\\
        Yelp18~\cite{wang2019neural} & 31,668 & 38,048 & 1,097,007 & NDCG@20\\
        \bottomrule
    \end{tabular}%
    
    \label{tab:datasets}
\end{table}
\subsection{Datasets}

We evaluate NBF-Rec on seven real-world recommendation datasets spanning movies (MovieLens-1M), music (LastFM), e-commerce (Amazon Beauty), location check-ins (Gowalla), product reviews (Epinions), and local business ratings (Yelp18). Each dataset is modeled as a bipartite user-item graph, with edge features such as ratings, timestamps, interaction counts, or metadata incorporated when available.
MovieLens-1M contains around 1M ratings from 5.9K users on 2.8K movies. LastFM includes implicit feedback (e.g., play counts) between 1.8K users and artists. Amazon Beauty and BookX represent large-scale product domains, with over 250K interactions each. Gowalla captures spatio-temporal check-ins, while Epinions includes explicit ratings and user trust relationships. Yelp18 contains over 1M user-business reviews with textual and temporal metadata.
Dataset statistics are summarized in Table~\ref{tab:datasets}.

\subsection{Experimental Setup}

For each dataset, we evaluate NBF-Rec in three settings:

\begin{itemize}
    \item \textbf{End-to-End Training}: The model is trained and tested on the same dataset.
    \item \textbf{Zero-Shot}: NBF-Rec is pretrained on one dataset and directly tested on another without fine-tuning. Embedding MLPs are randomly initialized. 
    \item \textbf{Fine-Tuning}: The whole pretrained model is fine-tuned on the target dataset.
\end{itemize}

We pretrain our models on datasets using both single-graph and multi-graph configurations, where we train on multiple datasets. Alongside our model, we also evaluate a simplified version of NBFNet that relies solely on structural graph information.
For evaluation, we use \textbf{Hits@K} and \textbf{NDCG@K}. Hits@K measures whether a relevant item appears in the top $K$ recommendations, while NDCG@K incorporates ranking information by normalizing the discounted cumulative gain (DCG) against the ideal DCG (IDCG). 
While Hits@K provides a binary assessment of whether a relevant item appears in the top-$K$ recommendations, NDCG@K captures both the presence and ranking of relevant items in the list.

\subsection{Results and Analysis}

\begin{figure}[t]
    \centering
    \includegraphics[width=0.8\linewidth]{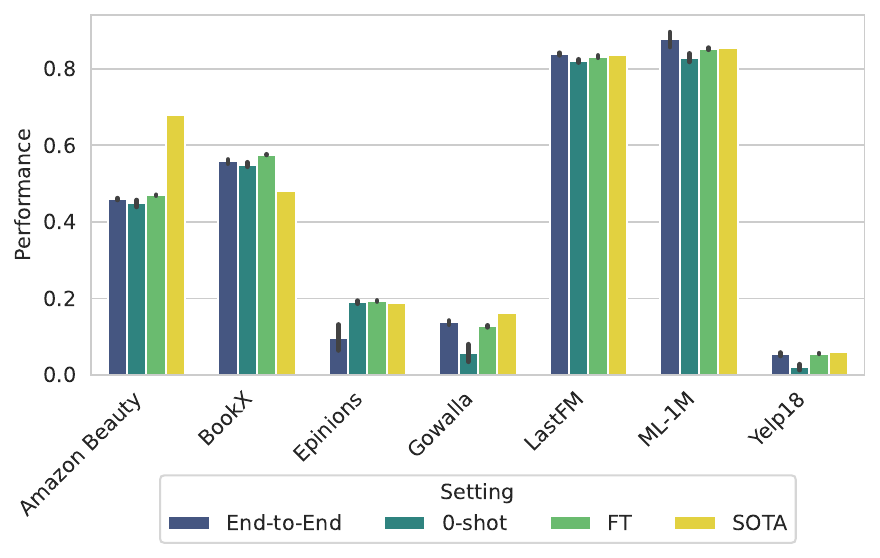}
    \caption{Performance of NBF-Rec in zero-shot, fine-tuned, and end-to-end settings across seven datasets. Metrics are Hits@10 or NDCG@20, depending on the dataset (see Table~\ref{tab:datasets}). Bars indicate 95\% confidence intervals over 3 runs. Zero-shot transfer achieves competitive results.}
    \label{fig:main_results}
\end{figure}

We evaluate the ability of NBF-Rec to generalize across recommendation domains in two inductive transfer settings: (i) zero-shot transfer, where no target domain data is seen during training, and (ii) fine-tuning, where the target data is used for supervised adaptation. Unless otherwise noted, results are derived from a model pretrained on Amazon Beauty and Epinions.
We report SOTA performance for each dataset based on results from prior work (ML-1M~\cite{rendle2012bpr}, LastFM~\cite{ning2011slim}, Amazon Beauty~\cite{lopez2024positional}, Gowalla~\cite{sun2023neighborhood}, Epinions~\cite{wang2019neural}, BookX~\cite{rendle2012bpr}, Yelp18~\cite{sun2023neighborhood}).
Figure~\ref{fig:main_results} shows the performance of NBF-Rec in these settings compared to a domain-specific model trained end-to-end on each dataset. We observe that zero-shot transfer yields competitive results across most datasets. For example, on MovieLens-1M, BookX, and Yelp18, the zero-shot model achieves performance within 5\% of the fully supervised baseline. This demonstrates that NBF-Rec learns representations and message-passing procedures that generalize to unseen user-item graphs without requiring retraining.
Fine-tuning the pretrained model on the target domain consistently improves performance across all datasets. The relative improvement is most pronounced on LastFM and Gowalla, where the zero-shot setting initially underperforms. These results suggest that even modest amounts of in-domain supervision suffice to close the gap to fully supervised models. Importantly, NBF-Rec supports this adaptation without reinitialization or retraining from scratch.

\begin{figure}[t]
    \centering
    \includegraphics[width=0.75\linewidth]{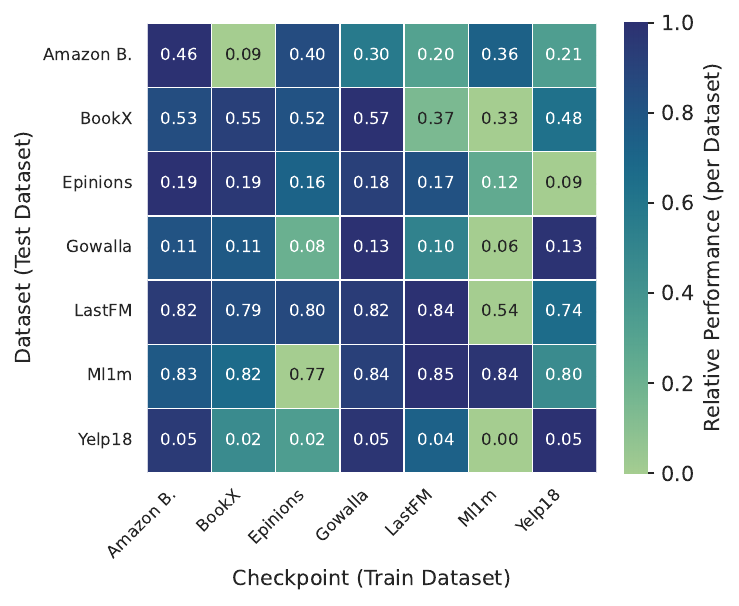}
    \caption{Cross-dataset transfer heatmap: each column indicates the training dataset and each row the test dataset. Cell colors represent relative performance (normalized by row). NBF-Rec achieves strong transfer when trained on Amazon Beauty, outperforming several dataset-specific models.}
    \label{fig:heatmap}
\end{figure}

\paragraph{\textbf{Cross-Dataset Transfer.}} We analyze dataset-to-dataset transferability based on the heatmap in Figure~\ref{fig:heatmap}. Along the diagonal, models are evaluated in a zero-shot setting using only the pretrained backbone checkpoint, without end-to-end training on the target domain. Notably, the diagonal does not always yield the best performance: for instance, BookX and Amazon Fashion achieve higher accuracy when transferred from other datasets than from their own backbone. This suggests that pretraining on different domains can provide more effective inductive biases than pretraining on the target domain itself. 
Edge features that are less informative, as observed in BookX and LastFM, generally reduce transferability. However, rich edge features alone do not guarantee successful transfer, as illustrated by Yelp’s relatively weak performance. Conversely, Gowalla, a large graph with fewer edge attributes, transfers effectively. Finally, transferability is not symmetric: for example, models trained on LastFM transfer well to MovieLens-1M, whereas the reverse transfer is less effective.

\paragraph{\textbf{Comparison with NBFNet.}}

We compare NBF-Rec to its base model, NBFNet, which solely relies on the graphs' structural information to assess the impact of our modifications (e.g., the inclusion of edge embeddings) on transferability and overall performance. 
Figure~\ref{fig:nbfnet-comparison} (Appendix) shows that in the zero-shot and fine-tuning settings, NBFRec consistently outperforms NBFNet. Even under end-to-end training, NBF-Rec usually performs on par with or slightly better than NBFNet.
These results confirm that NBF-Rec generalizes the NBFNet architecture by supporting inductive transfer with node and edge embeddings while maintaining strong in-domain performance.

\section{Conclusions}
We presented NBF-Rec, a graph-based recommender designed for inductive transfer learning across datasets with disjoint user and item sets. Unlike transductive models, NBF-Rec computes node representations dynamically and supports generalization without retraining.
Experiments on seven real-world datasets show that NBF-Rec performs competitively in zero-shot settings and improves further with fine-tuning, often matching end-to-end baselines. These results demonstrate the feasibility of transfer in graph-based recommendation without shared entities.
Our architecture primarily serves to demonstrate that inductive transfer is a viable approach for graph-based recommendation. While results vary across domains and our evaluation is limited to medium-scale datasets, the findings indicate that effective cross-domain generalization is achievable and highlight transfer learning as a promising direction for future recommender systems.

%
%
%
\bibliographystyle{unsrtnat}
\bibliography{bibliography}

\appendix
\section{Appendix}
\begin{figure}[hb]
    \centering
    \includegraphics[width=0.7\linewidth]{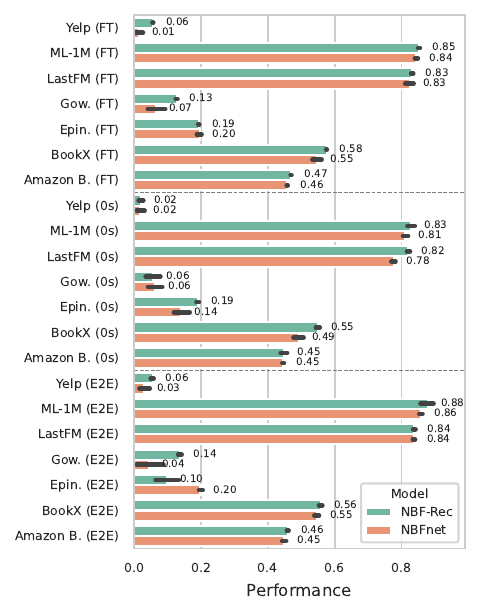}
    \caption{Performance comparison between NBF-Rec and NBFNet across all datasets under end-to-end, zero-shot, and fine-tuned settings. Incorporating interaction-level features and inductive training enables consistent improvements in recommendation quality.}
    \label{fig:nbfnet-comparison}
\end{figure}

We provide the comparison with NBFNet in Figure\ref{fig:nbfnet-comparison}.

\end{document}